%% file: main.tex
\documentclass[letterpaper, 10 pt, journal, twoside]{IEEEtran} 

\IEEEoverridecommandlockouts    
\usepackage{glossaries}
\input{glossary}

\usepackage{tabularx}
\usepackage{multirow, multicol, makecell}
\usepackage{array}
\newcolumntype{P}[1]{>{\centering\arraybackslash}p{#1}}
\newcolumntype{M}[1]{>{\centering\arraybackslash}m{#1}}
\usepackage{booktabs}
\newcolumntype{N}{>{\centering\arraybackslash}m{.5in}}
\newcolumntype{G}{>{\centering\arraybackslash}m{2in}}

\let\labelindent\relax

\usepackage[table,usenames,dvipsnames]{xcolor}      
\usepackage{extarrows}                              
\usepackage{enumitem}
\usepackage{wrapfig}
\usepackage{multirow} 
\usepackage{multicol} 
\usepackage{arydshln}
\usepackage{makecell}
\usepackage{dashrule}
\usepackage{threeparttable}
\usepackage{tablefootnote}
\usepackage{amsmath,amssymb,amsfonts,amsthm,dsfont, bm} 
\usepackage{mathtools}
\usepackage{algorithm,algorithmicx,listings}        
\usepackage[noend]{algpseudocode}			              
\definecolor{myBlue}{RGB}{0,0,0}
\definecolor{ver2blue}{RGB}{0,0,255}
\makeatletter
\def\BState{\State\hskip-\ALG@thistlm}
\makeatother
\DeclarePairedDelimiter\abs{\lvert}{\rvert}%
\DeclarePairedDelimiter\norm{\lVert}{\rVert}%
\makeatletter
\let\oldabs\abs
\def\abs{\@ifstar{\oldabs}{\oldabs*}}
\let\oldnorm\norm
\def\norm{\@ifstar{\oldnorm}{\oldnorm*}}
\makeatother

\usepackage{graphicx}
\usepackage{textcomp}
\usepackage[breaklinks=true, colorlinks, bookmarks=true, citecolor=Black, urlcolor=Violet,linkcolor=Black]{hyperref}

\usepackage[normalem]{ulem}
\usepackage[utf8]{inputenc}
\usepackage[english]{babel}
\usepackage{hyperref}
\hypersetup{
    colorlinks=true,
    linkcolor=blue,
    filecolor=magenta,      
    urlcolor=cyan,
}

\DeclareMathAlphabet\mathbfcal{OMS}{cmsy}{b}{n}

\usepackage[space, compress, sort]{cite}

\newtheorem*{assumption*}{Assumption}

\newtheorem*{problem*}{Problem}

\usepackage{lipsum}
\usepackage[capitalise]{cleveref}

\usepackage{tikz}

\crefname{section}{Sec.}{Secs.}
\Crefname{section}{Sec.}{Secs.}

\begin{document}

\title{{NEDS-SLAM: A Neural Explicit Dense Semantic SLAM Framework using 3D Gaussian Splatting}}


\author{Yiming Ji, Yang Liu$^{*}$, Guanghu Xie, Boyu Ma, Zongwu Xie, and Hong Liu %

\thanks{$^{*}$corresponding author}%
\thanks{All authors are with State Key Laboratory of Robotics and Systems, Harbin Institute of Technology. Email: yimingji@stu.hit.edu.cn (Yiming Ji), liuyanghit@hit.edu.cn (Yang Liu).
}}

\maketitle

\begin{abstract}

We propose NEDS-SLAM, \textcolor{myBlue}{a dense} semantic SLAM system based on 3D Gaussian representation, that enables robust 3D semantic mapping, accurate camera tracking, and high-quality rendering in real-time.
In the system, we propose a \textbf{Spatially Consistent Feature Fusion} model to reduce the effect of erroneous estimates from pre-trained segmentation head on semantic reconstruction, achieving robust 3D semantic Gaussian mapping.
Additionally, we employ a lightweight encoder-decoder to compress the high-dimensional semantic features into a compact 3D Gaussian representation, mitigating the burden of excessive memory consumption. Furthermore, we leverage the advantage of 3D Gaussian splatting, which enables efficient and differentiable novel view rendering, and propose a  Virtual Camera View Pruning method to eliminate \textcolor{myBlue}{outlier gaussians}, thereby effectively enhancing the quality of scene representations.
Our NEDS-SLAM method demonstrates competitive performance over existing dense semantic SLAM methods in terms of mapping and tracking accuracy on Replica and ScanNet datasets, while also showing excellent capabilities in 3D dense semantic mapping.

\begin{IEEEkeywords}
3D Gaussian Splatting; Dense Semantic Mapping; Neural SLAM; 3D Reconstruction.
\end{IEEEkeywords}

\end{abstract}
     \input{secs/introduction.tex}
     \input{secs/related_work.tex}
     \input{secs/method.tex}
     \input{secs/experiments.tex}
     \input{secs/conclusion.tex}
\bibliography{main}

\end{document}

%% file: glossary.tex
\newacronym{mav}{MAV}{Micro Aerial Vehicles}
\newacronym{uav}{UAV}{Unmanned Aerial Vehicle}
\newacronym{ovc}{OVC}{Open Vision Computer}
\newacronym{lidar}{LiDAR}{Light Detection and Ranging}
\newacronym{vio}{VIO}{Visual-Inertial Odometry}
\newacronym{gpgpu}{GPGPU}{General-Purpose Graphics Processing Unit}
\newacronym{ugv}{UGV}{Unmanned Ground Vehicle}
\newacronym{uwb}{UWB}{Ultra Wideband}
\newacronym{svm}{SVM}{Support Vector Machine}
\newacronym{fcn}{FCN}{Fully Convolutional Network}
\newacronym{cnn}{CNN}{Convolutional Neural Network}
\newacronym{loam}{LOAM}{LiDAR Odometry and Mapping}
\newacronym{sloam}{SLOAM}{Semantic LiDAR Odometry and Mapping}
\newacronym{slam}{SLAM}{Simultaneous Localization and Mapping}
\newacronym{iot4ag}{IoT4Ag}{NSF Engineering Research Center for the Internet of Things for Precision Agriculture}
\newacronym{grasp-lab}{GRASP Lab}{the General Robotics, Automation, Sensing and Perception Laboratory}
\newacronym{jps}{JPS}{Jump Point Search}
\newacronym{ukf}{UKF}{Unscented Kalman Filter}
\newacronym{sam}{SAM}{Smoothing and Mapping}
\newacronym{slc}{SLC}{Semantic Loop Closure}
\newacronym{icp}{ICP}{Iterative Closest Point}
\newacronym{imu}{IMU}{Inertial Measurement Unit}
\newacronym{tsdf}{TSDF}{Truncated Signed Distance Field}
\newacronym{esdf}{ESDF}{Euclidean Signed Distance Field}
\newacronym{sdf}{SDF}{Signed Distance Field}
\newacronym{rrt}{RRT}{Rapidly Exploring Random Tree}
\newacronym{fpv}{FPV}{First-person View}
\newacronym{dnn}{DNN}{Deep Neural Network}
\newacronym{igpred}{IGPred}{Information Gain Prediction}
\newacronym{csqmi}{CSQMI}{Cauchy-Schwarz Quadratic Mutual Information}
\newacronym{nbv}{NBV}{Next Best View}
\newacronym{vae}{VAE}{Variational Autoencoder}
\newacronym{tsp}{TSP}{Travelling Salesman Problem}
\newacronym{bgsm}{BGSM}{Behavior Guidance State Machine}
\newacronym{pca}{PCA}{Principal Component Analysis}
\newacronym{aspp}{ASPP}{Atrous Spatial Pyramid Pooling}
\newacronym{swap}{SWaP}{Size Weight and Power}
\newacronym{soi}{SoI}{Semantic Object of Interest}
\newacronym{aoi}{AoI}{Area of Interest}
\newacronym{cop}{COP}{Correlated Orienteering Problem}
\newacronym{ig}{IG}{Information Gain}

%% file: secs/introduction.tex
\section{Introduction}
\label{sec:intro}
Visual SLAM (Simultaneous Localization and Mapping) is a fundamental research problem in robotics, which involves simultaneously tracking the camera pose and incrementally constructing a map of an unknown environment \cite{newcombe2011kinectfusion}.
Downstream tasks such as autonomous goal navigation, human-computer interaction, mixed reality (MR), and augmented reality (AR) demand not only accurate camera pose tracking from SLAM systems but also robust and dense semantic reconstruction of the environment.
This research focuses on semantic RGBD-SLAM, which, in contrast to traditional SLAM, enables the identification, classification, and association of entities within a scene, ultimately generating a semantically-rich map.

Inspired by the success of NERF and 3D Gaussian Splatting (3DGS) in high-fidelity view synthesis, researchers have explored building end-to-end visual SLAM systems based on neural radiance fields. These novel SLAM architectures offer superior solutions compared to traditional algorithms in terms of surface continuity, memory requirements, and scene completion.
Specifically, iMAP \cite{sucar2021imap} and NICE-SLAM \cite{zhu2022nice} leverage neural implicit fields for consistent geometry representation, while \textcolor{myBlue}{MonoGS \cite{Matsuki:Murai:etal:CVPR2024}} and \textcolor{myBlue}{SplaTAM \cite{keetha2023splatam} employ 3DGS} to achieve photo-realistic mapping.


\textcolor{myBlue}{
Given continuous input of RGB-D frames, dense semantic SLAM aims to create a compact and dense 3D representation of the scene that includes accurate RGB information as well as dense semantic data.
However, current state-of-the-art semantic segmentation models are trained on large amounts of internet images, which are loosely related and time-independent. This leads to estimation errors such as semantic spatial inconsistency, which significantly impairs the density and completeness of semantic reconstruction.
The previous 3DGS-based semantic SLAM method \cite{li2024sgs} overlooked the issue of semantic feature inconsistency, which limits its potential for practical applications.
}

\textcolor{myBlue}{
Furthermore, our research has found that directly embedding semantic category labels into gaussians parameters may not be appropriate.
During splatting, overlapping gaussians combine through alpha-blending to form pixel values on the imaging plane.
Using RGB color channels as an example, ideally, when 3D gaussians are splatted onto different imaging planes, they create different color blending effects. However, assigning fixed class labels to the gaussians leads to meaningless values in the semantic channels during splatting.
Therefore, attempting to embed semantic features instead of semantic category labels into the 3D gaussians parameters would be more promising. However, this approach can cause prohibitive memory requirements and significantly lower the efficiency of both optimization and rendering, as semantic features typically have higher dimensions, whereas category labels are just integer values.
}

\textcolor{myBlue}
{
    In a 3DGS-based SLAM system, the process of incrementally building a map is often influenced by camera pose estimation errors, object occlusions, and errors in the optimization process. These factors can introduce 3D gaussians that do not align with actual surfaces. When these outlier gaussians are included in the rendering view, they can create visual artifacts, which in turn affect camera pose estimation, creating a vicious cycle. This issue is not addressed in the original 3DGS paper, where the camera poses for each frame are precomputed using an offline SFM method. Therefore, handling outlier gaussians is crucial for 3DGS-based SLAM methods.
}

Overall, \textcolor{myBlue}{3DGS based Dense Semantic SLAM} can be summarized as facing two key challenges: 1) Providing robust semantic reconstruction results under inconsistent semantic features. 2) Incrementally building a map that can accurately distinguish well-optimized and low-quality regions, while effectively filtering out outliers to improve reconstruction quality.

This paper proposes NEDS-SLAM, with the following key contributions:

\begin{itemize}

\item 
{\textcolor{myBlue}{We propose the Spatially Consistent Feature Fusion module (SCFF), which combines semantic features with appearance features. This module addresses the spatial inconsistency of semantic features and provides a more robust semantic SLAM solution.}}
\item 
{\textcolor{myBlue}{We embed semantic features into Gaussian parameters instead of using category labels. We also introduce a lightweight encoder-decoder to prevent memory issues from high-dimensional semantic feature embedding.}}
\item {\textcolor{myBlue}{We present the Virtual Camera View Pruning (VCVP) method. VCVP generates multiple virtual camera views to ensure consistency, identifying and removing unstable gaussians caused by occlusions, camera pose errors, and parameter optimization issues, leading to a more accurate 3D Gaussian field.}}
\end{itemize}

%% file: secs/related_work.tex
\section{Related Work}

\subsection{Traditional approaches to dense semantic SLAM}
Real-time dense semantic SLAM systems face the challenge of effectively fusing semantic information into underlying 3D geometric representations of the environment. Traditional approaches use voxels, point clouds, and signed distance fields to encode object labels \cite{hermans2014dense_vox},\cite{narita2019panopticfusion_sdf}.
However, voxel- and point cloud-based approaches struggle with reconstruction speed and high-fidelity model acquisition. Meanwhile, signed distance field representations incur high memory usage that does not scale well to large-scale environments. There remains a need for more efficient and expressive 3D semantic modeling techniques suitable for real-time dense SLAM.

\textcolor{myBlue}{
\subsection{NeRF based SLAM}
In recent years, Neural Radiance Fields (NeRF) have sparked significant interest in computer graphics, attracting attention for their high-fidelity novel view synthesis and lightweight scene representation \cite{mildenhall2021nerf}. This enthusiasm has quickly spread to the SLAM field, leading to the development of many innovative SLAM architectures \cite{sucar2021imap}\cite{zhu2022nice}.
Zhu et al. introduced SNI-SLAM \cite{zhu2023sni}, which employs neural implicit representation and hierarchical semantic encoding for multi-level scene understanding, contributing a cross-attention mechanism for the collaborative integration of appearance, geometry, and semantic features.
Due to the limitations of NeRF's volume rendering, NeRF-based dense semantic SLAM struggles to simultaneously model and optimize the semantic and RGB-geometry information of the environment \cite{li2023dns}\cite{haghighi2023neural}. Additionally, the efficiency of SLAM is constrained by the implicit representation of the map\cite{tosi2024nerfs}.
}

\subsection{Gaussian Splatting based SLAM}
\textcolor{myBlue}{3DGS} representations have emerged as a promising approach for 3D scene modelling using a set of 3D \textcolor{myBlue}{gaussians}, each characterized by parameters such as position, anisotropic covariance, opacity, and color \cite{kerbl20233d}.
While existing \textcolor{myBlue}{3DGS-based} SLAM methods have primarily focused on RGB reconstruction, exploring end-to-end system architectures, optimization of \textcolor{myBlue}{gaussians} parameters, and accurate camera pose tracking through differentiable rendering, less attention has been paid to semantic reconstruction \cite{keetha2023splatam},\cite{yan2023gs},\cite{Matsuki:Murai:etal:CVPR2024},\cite{yugay2023gaussian}.
The few semantic \textcolor{myBlue}{3DGS-SLAM} approaches proposed to date have simply encoded ground truth semantic color labels directly as a second color channel of the \textcolor{myBlue}{gaussians} parameters \cite{li2024sgs}, without explicit modeling of semantic information or inference. There is clear potential for more sophisticated integration of semantics within the \textcolor{myBlue}{3DGS-SLAM} framework.
The present work conducts a more in-depth exploration of \textcolor{myBlue}{dense semantic} SLAM, aiming to simultaneously improve the robustness and reconstruction fidelity of \textcolor{myBlue}{3DGS-based} SLAM systems through more sophisticated modeling and inference of semantic information within the \textcolor{myBlue}{3DGS} representation.

%% file: secs/method.tex
\begin{figure*}
    \begin{center}
        \includegraphics[width=1\linewidth]{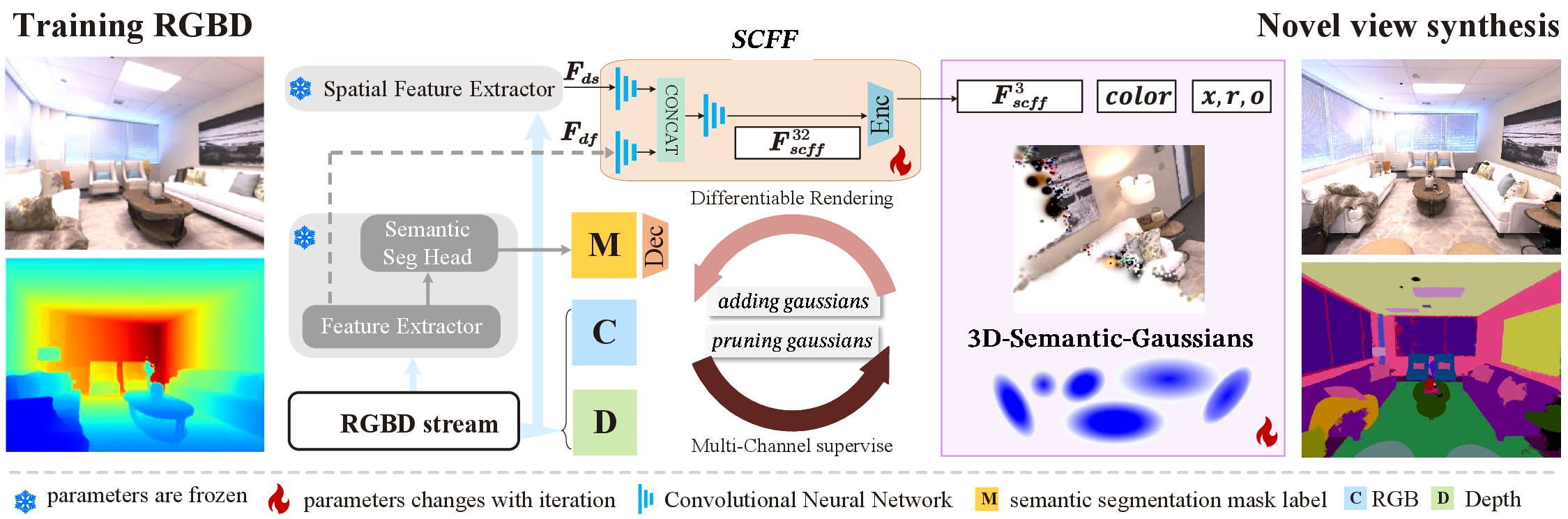}
    \end{center}
    \caption{
        Overview of the proposed NEDS-SLAM. Our method takes an RGB-D stream as input. RGB images are processed by the pretrained semantic feature extractor to get semantic features, while dense appearance features are obtained through the Spatial Feature Extractor model.
        The semantic and appearance features are fused to generate high-dimensional semantic features that are spatially consistent. These features are then processed by the encoder to generate low-dimensional features and embedded into the GS parameters.
        By employing Differentiable Rendering, real RGB images, depth images, and semantic masks predicted by a pre-trained segmentation head are utilized for Multi-Channel supervision. This approach enables the joint optimization of GS parameters.
        In the figure, $M$, $C$, and $D$ represent the semantic segmentation mask, color, and depth information, respectively.
        NEDS-SLAM achieves high-fidelity map reconstructions while simultaneously accomplishing compact and dense pixel-level semantic reconstruction.
        }
    \label{fig:fig1}
\end{figure*}
\section{Methodology}
\label{sec:method}

\subsection{Scene Representation and Semantic embedding}
\label{sec:3D splatting}
Each 3DGS utilized for representing three-dimensional scenes encompasses mean, covariance, and color information. In this paper, a simplified 3DGS representation of the scene is employed \cite{keetha2023splatam}, omitting the spherical harmonics functions used for color representation, while assuming \textcolor{myBlue}{gaussians} to be isotropic \textcolor{myBlue}{as in Eq \ref{eq:eq1}}.

\begin{equation}
    \label{eq:eq1}
    f^{gs}\left( \bm{x} \right) = o\exp \left( { - \frac{{{{\left\| {\bm{x} - \mu } \right\|}^2}}}{{2{r^2}}}} \right)
\end{equation}

Where $\mu \in \mathbb{R}^3$ \textcolor{myBlue}{represents the center position}, $r$ is the radius, and $o \in [0,1]$ represents the opacity.
The rapid and differentiable rendering based on \textcolor{myBlue}{3DGS} serves as the core of mapping and tracking within \textcolor{myBlue}{3DGS-based} SLAM systems. This ability for fast rendering enables the system to directly compute the gradients of the underlying parameters based on the discrepancy between the rendered results and the actual data. Consequently, the \textcolor{myBlue}{gaussians} parameters can be updated to achieve an accurate representation of the scene.
The differentiable rendering process based on \textcolor{myBlue}{gaussians splatting} comprises three steps: Frustum Culling, Splatting, and Rendering by Pixels \cite{chen2024survey2}. 
\begin{equation}
    \label{eq:eq2}
    C\left( p \right) = \sum\limits_{i \in N} {\bm{c}_i} f_i^{gs}\left( p \right)\prod\limits_{j = 1}^{i - 1} {\left( {1 - f_j^{gs}\left( p \right)} \right)} 
\end{equation}

After arranging a collection of 3D \textcolor{myBlue}{gaussians} and camera pose, it is imperative to sort the \textcolor{myBlue}{gaussians} in a front-to-back manner. By employing alpha-compositing, the splatted 2D projection of each \textcolor{myBlue}{gaussian} can be efficiently rendered in pixel space, ensuring the generation of RGB images in the desired order, as Eq \ref{eq:eq2}. 
$\bm{c}_i$ represents the color parameters of the \textcolor{myBlue}{gaussians}, and $f_i^{gs}\left( p \right)$ is computed as in Eq \ref{eq:eq1} but with the 2D splatted ${\mu}$ and ${r}$. The rendering process is completed by multiplying the opacity of each \textcolor{myBlue}{gaussian} with the color and accumulating the results. The depth map is rendered in a similar manner, as shown in Eq \ref{eq:eq3}.

\begin{equation}
    \label{eq:eq3}
    D\left( p \right) = \sum\limits_{i \in N} {\bm{d}_i} f_i^{gs}\left( p \right)\prod\limits_{j = 1}^{i - 1} {\left( {1 - f_j^{gs}\left( p \right)} \right)} 
\end{equation}

The most notable distinction between semantic features and color and geometric features lies in their high-dimensional attributes. 
The semantic features do not refer to the per-pixel class labels generated by the segmentation head. Instead, it pertains to the high-dimensional semantic features extracted by the pre-trained model at each pixel.
Taking DINO \cite{oquab2023dinov2} as an example, the ViT-S model produces latent feature encodings of 384 dimensions, while the ViT-G model produces encodings of 1536 dimensions.

A simple way to combine \textcolor{myBlue}{3DGS} with semantic features is to add trainable feature vectors to each \textcolor{myBlue}{gaussian}. These parameters can be learned during the differentiable rendering process, which allows end-to-end training.
However, for \textcolor{myBlue}{dense semantic} SLAM, adding a high dimensional semantic feature vector to each 3DGS is memory-inefficient. 
Inspired by LangSplat \cite{qin2023langsplat}, we propose using a simple MLP as an encoder to compact semantic features into a low-dimensional vector. The compressed semantic features are then added to the 3D gaussians and can be rendered as in Eq \ref{eq:eq4}.
\begin{equation}
    \label{eq:eq4}
    S\left( p \right) = \sum\limits_{i \in N} {\bm{f}_i} f_i^{gs}\left( p \right)\prod\limits_{j = 1}^{i - 1} {\left( {1 - f_j^{gs}\left( p \right)} \right)} 
\end{equation}

\subsection{Adaptive 3D Gaussian Expansion Mapping}
\subsubsection{\textbf{Spatially Consistent Feature Fusion \textcolor{myBlue}{(SCFF)}} }
\textcolor{myBlue}{
Pervious semantic SLAM approaches typically use pretrained segmentation models to compute pixel-level labels from each RGB frame,but these class labels lack environmental specificity. Pretrained models may produce inconsistent semantic estimates, where the same object is predicted with different semantic labels in images from different camera views.}


To address this issue, SNI-SLAM \cite{zhu2023sni} computes a fused feature by combining geometry, appearance, and semantic features.
CoSSegGaussians \cite{dou2024cosseggaussians} incorporates DINO \cite{oquab2023dinov2} features with superior multi-view semantic scale consistency into the \textcolor{myBlue}{gaussians} parameters. Subsequently, the semantic encoding of each \textcolor{myBlue}{gaussians} is fused with spatial coordinates to render semantic features, thereby enhancing robustness.

In this paper, we propose a simplified fusion mechanism. \textcolor{myBlue}{It combines the appearance features with the semantic features extracted from pretrained model}. The resulting mixed feature, obtained through MLP \textcolor{myBlue}{encoder}, is then embedded as the final semantic encoding in the 3DGS representations.

\textcolor{myBlue}{
    As shown in Fig \ref{fig:fig1}, the pretrained semantic feature extractor extracts an {$H \times W \times D_f$} feature map $F_{df}$ from an {$H \times W \times 3$} RGB frame.
    At the same time, the spatial feature extractor extracts {$H \times W \times D_s$} features $F_{ds}$ from RGB data.
    After three layers of convolution, the feature channels of $F_{df}$ are reduced to 256, 128, and 16, respectively. Similarly, the feature channels of $F_{ds}$ are increased to 16 through one layer of CNN. 
    After concatenation and the final convolution, we obtain the spatially consistent feature $F^{32}_{scff}$ with 32 channels.
    Using the external parameters of the camera, we can convert an input frame of RGBD into a series of points in 3D space. Each point includes $xyz$ coordinates, RGB information, and 32-channel SCFF features.
    }

\textcolor{myBlue}{
    To reduce the number of 3D gaussian parameters, we need to use an information encoding method to compress $F^{32}_{scff}$ to a lower dimension, such as using hash encoding \cite{zuo2024fmgs}, GPR \cite{yuan2024uni}, etc.
    In this paper, we use a simple MLP to compress $F^{32}_{scff}$ to three dimensions, resulting in $F^{3}_{scff}$.
} 

\textcolor{myBlue}{
    It is important to clarify that the semantic category labels are numerical IDs from a predefined category library (e.g., 0 represents a person), while the $F^{3}_{scff}$ values range between 0 and 1. These semantic features can be decoded back into semantic category labels by a subsequent decoder.
}

\textcolor{myBlue}{
    We use the pre-trained DINO \cite{oquab2023dinov2} model as a semantic feature extractor, obtaining features $F_{df}$ with 384 channels ($D_f=384$).
    We use DepthAnything \cite{yang2024depth} as the spatial feature extractor, resulting in $D_s=1$.
    Relative depth output from DepthAnything is used as the appearance feature because changes in the camera viewpoint do not affect the relative position of surfaces on the object.
    The spatial consistency of appearance features helps SCFF achieve stable semantic feature estimation.
}



The relative depth between pixels can reflect the geometric structure of observed surfaces. The \textcolor{myBlue}{SCFF module} dynamically adjusts the weights of semantic features according to the spatially consistent relationships. It thereby reduces the impact of segmentation errors on the spatial consistency of semantic features.
\subsubsection{\textbf{Updating 3D Gaussians}}
During the mapping process, we assume that the camera pose for the current frame is known. We need to use the current keyframe's RGBD data to update the \textcolor{myBlue}{gaussians representation} of the scene. \textcolor{myBlue}{Updating has two meanings: optimizing existing scene parameters and generating a new 3DGS distribution for the scene}.

Following the processes used in Splatam \cite{keetha2023splatam} and GS-SLAM \cite{yan2023gs}, \textcolor{myBlue}{we use Eq.\ref{eq:sil_compute} to calculate the silhouette value per pixel}. The silhouette images are rendered to determine the contribution of each \textcolor{myBlue}{gaussian} to the map.

\begin{equation}
    \textcolor{myBlue}{
    \label{eq:sil_compute}
    Sil\left( p \right) = \sum\limits_{i \in N} f_i^{gs}\left( p \right)\prod\limits_{j = 1}^{i - 1} {\left( {1 - f_j^{gs}\left( p \right)} \right)} 
    }
\end{equation}

At the same time, the difference between the projected depth value and the ground truth value of pixels corresponding to newly added \textcolor{myBlue}{gaussian} is checked when they are projected back onto the image plane.
\begin{equation}
    \label{eq:eq5}
    M\left( p \right) = \left[ {Sil\left( p \right) < {T_s}} \right] + \left[ {\left( {{D_{gt}}\left( p \right) - D\left( p \right)} \right) < {T_d}} \right]
\end{equation}

The densification mask $M\left( p \right)$ is calculated according to Eq \ref{eq:eq5}, where $D\left( p \right)$ represents the depth value of pixel $p$.
\textcolor{myBlue}{$M\left( p \right)$ represents a Boolean mask for pixel $p$. The optimization of 3DGS and the addition of new gaussians will be confined to areas where the mask value is True, thereby avoiding the densification of gaussians in well-reconstructed areas. This differs from the approach in \cite{kerbl20233d}, which splits gaussians in over-reconstructed regions. Due to the high real-time requirements of SLAM systems, setting threshold parameters in $M\left( p \right)$ allows the system to avoid the heavy computation associated with the gaussians densification method in \cite{kerbl20233d}.}


After the process discussed in Section \ref{sec:3D splatting}, the scene representations contains three feature channels: spatial position, surface color, and potential semantics. The spatial position and surface color are directly obtained from the RGBD data stream. Meanwhile, the fusion of semantic encoding is supervised by the mask output from a pretrained segmentation model.
\begin{equation}
    \label{eq:eq6}
    {L_c} = \lambda {L_1}\left( {{I_{r}},{I_{gt}}} \right) + \left( {1 - \lambda } \right)\left[ {1 - ssim\left( {{I_{r}},{I_{gt}}} \right)} \right]
\end{equation}
The color loss $L_c$ is represented as a weighted combination of SSIM \cite{kerbl20233d} and $L1$ loss as in Eq \ref{eq:eq6}. 
\begin{equation}
    \label{eq:eq7}
    {L_d} = \sum\limits_{pix} {\left| {D_{pix}^{render} - D_{pix}^{gt}} \right|}
\end{equation}
The depth loss $L_d$ is calculated as in Eq \ref{eq:eq7}. During the mapping stage, the multi-channel loss is as shown in Eq \ref{eq:eq8}, 
where $S_{render}$ represents the \textcolor{myBlue}{semantic labels after decoding the semantic features} and $S_{head}$ represents the \textcolor{myBlue}{class labels} computed by the pretrained model. \textcolor{myBlue}{We use the cross-entropy loss ${L_{CE}}$ to supervise the semantic channel}.
\begin{equation}
    \label{eq:eq8}
    {L_{mapping}} = {\lambda _c}{L_c} + {\lambda _d}{L_d} + {\lambda _s}{L_{CE}}\left( {{S_{render}},{S_{head}}} \right)
\end{equation}
In Eq \ref{eq:eq8}, $\lambda _d$, $\lambda _s$, and $\lambda _c$ are predefined hyperparameters used to assign weighted values to the depth, semantic, and color channels respectively.

\subsubsection{\textbf{\textcolor{myBlue}{Vitrual Camera View Pruning 3D Gaussians (VCVP)}}}
The key aspects of GS-based SLAM are: 1) Distinguishing well-established areas from areas requiring further optimization, and 2) Identifying and removing outlier points. The former resolves where to add \textcolor{myBlue}{gaussians}, and also plays a key role in camera tracking. Areas of low quality can severely affect the accuracy of pose tracking. The second key aspect resolves where to delete \textcolor{myBlue}{gaussians}. Outlier points will cause holes and defects during image rendering, and these flaws can also affect the accuracy of camera tracking.

The distinction between well-optimized and areas with low quality is implemented through Eq \ref{eq:eq5}. This section discusses issues related to \textcolor{myBlue}{gaussians} pruning.

\textcolor{myBlue}{
    Multi-view consistency constraints have been proven effective in identifying geometrically unstable gaussians. Previous methods \cite{Matsuki:Murai:etal:CVPR2024} check whether gaussians inserted within the latest three frames of a keyframe window are recorded by other keyframes, thereby determining outlier gaussians. 
    This method improves mapping accuracy by using collaborative constraints among multiple keyframes, but it increases computational costs and reduces real-time performance. Drastic viewpoint changes during SLAM cause significant overlap variations between keyframes, leading to errors in outlier detection.
    }

\begin{figure}
    \centering 
    \includegraphics[width=6 cm]{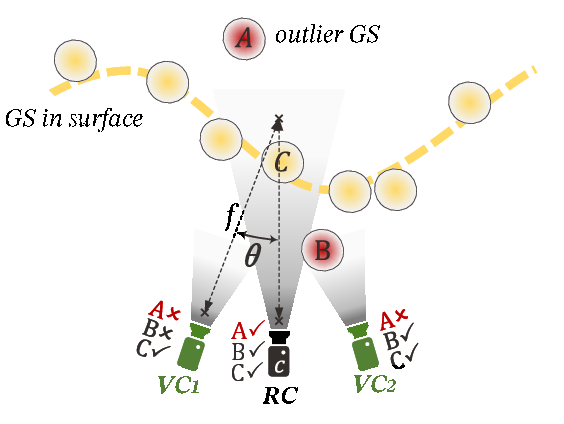}
    \caption{
        The concept of virtual view pruning for identifying outlier gaussians. We analyze only the gaussians visible in the current ground-truth view (points $A$, $B$, $C$ in the figure).
        Point A is not visible from either of the two virtual views, thus identified as an outlier gaussians, and its opacity is degraded during subsequent optimization. While the figure depicts two virtual views in a planar scenario, our approach creates four virtual cameras by rotating the camera pose from the focal point of each GT view frame along four directions: up, down, left, and right.
    }
    \label{fig:fig3}
\end{figure}
\begin{figure}
    \centering 
    \includegraphics[width=0.99\linewidth]{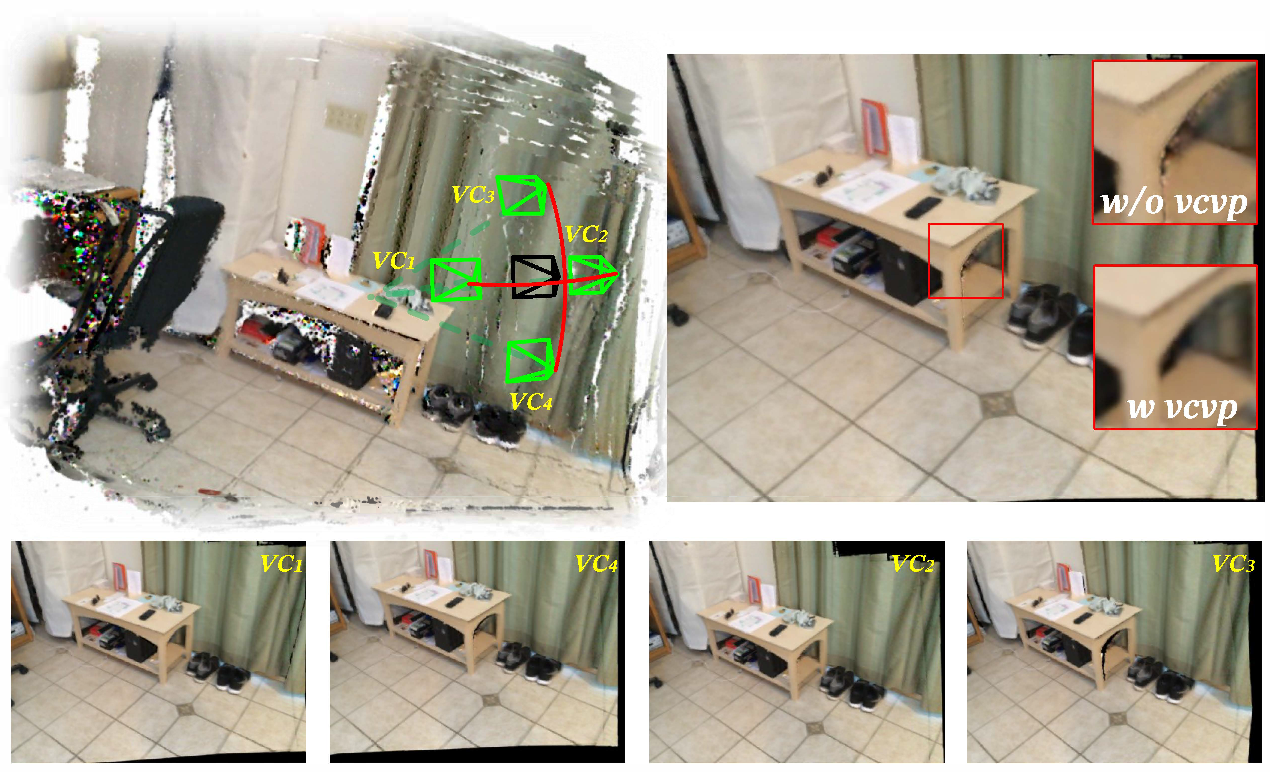}
    \caption{
        Rendered virtual camera views on the ScanNet dataset.
        The middle images provide a zoomed-in illustration of the effectiveness of Virtual Camera Pruning, where 'vcvp' denotes virtual camera view. Eliminating outlier \textcolor{myBlue}{gaussians} not only improves rendering quality but also reduces the storage footprint of the map representation.
        }
    \label{fig:vvpruning}
\end{figure}
\textcolor{myBlue}{
In contrast to this method, the VCVP method proposed in this paper does not perform comparison between keyframes. Instead, it compares the viewpoint between a real camera frame and the corresponding virtual accompanying camera frame, as depicted in Fig \ref{fig:fig3}. 
The virtual cameras ($VC$) are created by rotating the real camera {$\pm \theta$} around the focal point: $VC_1$ and $VC_2$ by rotating on the horizontal plane ($xz$ plane), and $VC_3$ and $VC_4$ by rotating on the vertical plane ($yz$ plane).}

The points $A$ and $B$ represent \textcolor{myBlue}{outlier gaussians}, while the GT view denotes the camera pose estimated within the RGBD stream. In the current keyframe, both $A$ and $B$ are visible. However, in the \textcolor{myBlue}{$VC_1$}, neither of these outlier points is visible, and in the \textcolor{myBlue}{$VC_2$}, only $B$ is visible while $A$ is not. The virtual camera operates alongside the real camera. If a \textcolor{myBlue}{gaussians} is invisible in all virtual views but visible in the real view, it is then considered an outlier.

The virtual multi-view consistency check method takes advantage of the fast rendering capabilities of the Gaussian Splatting, enabling the marking of \textcolor{myBlue}{gaussians} that significantly deviate from the object surface. 
\textcolor{myBlue}{
The VCVP method eliminates the dependence on historical keyframes, allowing it to remain unaffected by drastic changes in camera views. This enables the identification of single-view outlier gaussians.
}
In subsequent optimization processes, the involvement of outlier \textcolor{myBlue}{gaussians} in the scene is diminished by degrading their opacity. Consistent with \cite{kerbl20233d}, \textcolor{myBlue}{gaussians} with near-zero opacity or excessive radius are removed in the mapping process.
As illustrated in Fig \ref{fig:vvpruning}, we render virtual views and further optimize the \textcolor{myBlue}{3D gaussians parameters} only for keyframes. The specific approach for generating virtual views is not fixed. Although Gaussian splatting enables extremely fast virtual view synthesis (nearly 300 FPS), introducing too many viewpoints can compromise the system's real-time performance. 
\textcolor{myBlue}{
We conducted detailed tests in Section \ref{sec:ablation} to evaluate how the generation and function of the virtual camera impact the performance of the SLAM system.
}We choose four virtual views along the up, down, left, and right directions, which achieves a desirable balance between effectiveness and efficiency.

\subsubsection{\textbf{Camera tracking}}
The camera tracking phase involves estimating the relative pose of the camera for each new frame, based on the already established map model. The camera pose for the new frame is initialized under the assumption of constant velocity, which includes both a constant linear and angular velocity.

The camera pose is subsequently refined iteratively by minimizing the tracking loss between the ground truth of the color, depth, and semantic channels and the gaussian rendered results from the camera's perspective.
\begin{equation}
    \label{eq:eq10}
        L_{\text{tracking}} = \left( \lambda_c L_c + \lambda_d L_d + \lambda_s L_{\text{CE}}\left( S_{\text{render}}, S_{\text{head}} \right) \right) \cdot M
\end{equation}
$M$ in Eq \ref{eq:eq10} is computed as Eq \ref{eq:eq5}.  
Artifacts and flaws such as holes and spurious effects caused by outlier gaussians significantly impact the precision of camera tracking. Subsequent experiments demonstrate that the incorporation of semantic loss improve the tracking accuracy. This improvement is attributed to the enriched understanding of the geometric information of objects, facilitated by the integration of semantic features.

%% file: secs/experiments.tex
\begin{figure}
    \begin{center}
        \includegraphics[width=1\linewidth]{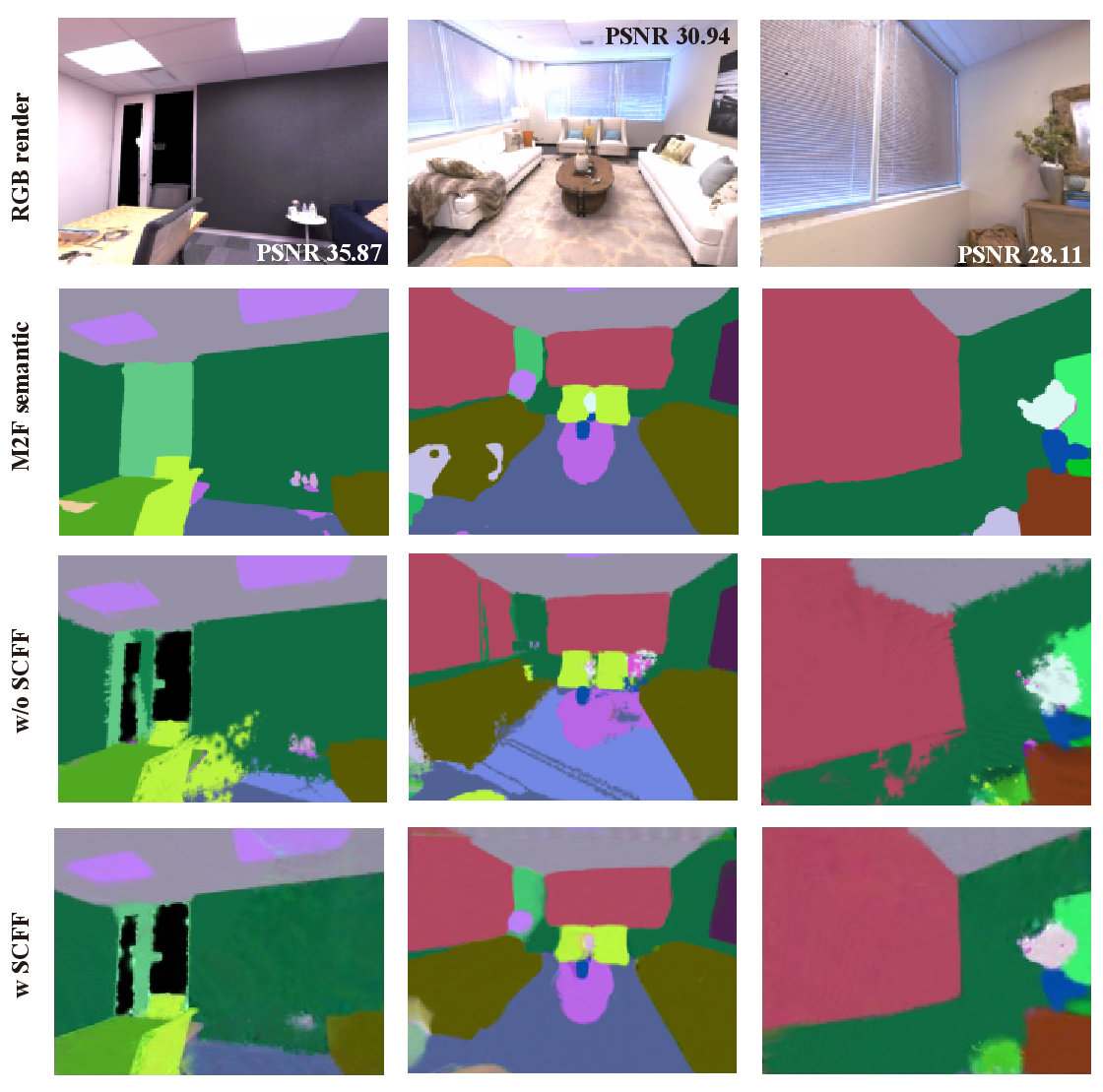}
    \end{center}
    \caption{
        The first row shows the RGB reconstruction results. The second row shows the semantic labels predicted directly on the current frame using M2F\cite{cheng2021mask2former}. The third row shows the semantic reconstruction results using the SGS-SLAM\cite{li2024sgs} method based on SplaTAM\cite{keetha2023splatam}. The fourth row shows the reconstruction results of our proposed model.
        }
    \label{fig:fig4}
\end{figure}

\begin{table}
    \centering
    \caption{
        Comparison experiments with other methods on map reconstruction and localization accuracy}
    \resizebox{\linewidth}{!}{%
    \begin{tabular}{l:cccc:c} 
    \toprule
    \multicolumn{1}{c}{Methods}                       & Depth L1[cm]$\downarrow$  & LPIPS$\downarrow$ & SSIM$\uparrow$ & \multicolumn{1}{c}{PSNR$\uparrow$} & ATE RMSE[cm] $\downarrow$    \\ 
    \midrule
    NICE-SLAM~\cite{zhu2022nice}     & 1.903         & 0.23              & 0.81           & 24.22                              & 2.503           \\
    Vox-Fusion~\cite{yang2022vox}    & 2.913         & 0.24              & 0.80           & 24.41                              & 1.473           \\
    Co-SLAM~\cite{wang2023co}        & 1.513         & 0.336             & 0.94           & 30.24                              & 1.059           \\
    ESLAM~\cite{johari2023eslam}     & 0.945         & 0.34              & \textbf{0.929} & 29.08                              & 0.678           \\
    SplaTAM \cite{keetha2023splatam} & \uline{0.49}  & \uline{0.10}      & 0.97           & \uline{34.11}                      & \uline{0.36}    \\
    NEDS-SLAM(Ours)                                   & \textbf{0.47} & \textbf{0.088}    & \uline{0.962}  & \textbf{34.76}                     & \textbf{0.354}  \\
    \bottomrule
    \end{tabular}
    }
    \label{tab:exp1}    
\end{table}

\begin{table}
    \centering
    \caption{Comparison experiment on the ATE RMSE metric}
    \resizebox{\linewidth}{!}{%
    \begin{tabular}{l:cccc:c} 
    \toprule
    \multicolumn{1}{c}{Methods} & scene0000      & scene0169      & scene0181      & \multicolumn{1}{c}{scene0207} & Avg.            \\ 
    \midrule
    NICE-SLAM\cite{zhu2022nice}                   & 12.00          & 10.90          & 13.40          & 6.20                          & 10.63           \\
    Vox-Fusion\cite{yang2022vox}                  & 68.84          & 27.28          & 23.30          & 9.41                          & 32.21           \\
    Point-SLAM\cite{sandstrom2023point}                  & \textbf{10.24} & 22.16          & 14.77          & 9.54                          & 14.18           \\
    SplaTAM\cite{keetha2023splatam}                    & 12.56          & \textbf{11.09} & \uline{11.07}  & \uline{7.46}                  & \uline{10.54}   \\ 
    NEDS-SLAM(Ours)             & \uline{12.34}  & \uline{11.21}  & \textbf{10.35} & \textbf{6.56}                 & \textbf{10.12}  \\
    \bottomrule
    \end{tabular}
    }
    \label{tab:exp_scannet}
    \end{table}

\section{Experiment}
\label{sec:experiment}
\subsection{Experimental Setup}
\label{sec:experiment_setup}
\noindent\textbf{Dataset.} 
We evaluate our method on both synthetic and realworld datasets with semantic maps. Following other nerf-based and gaussian-based SLAM methods, for the reconstruction quality, we evaluate quantitatively on 8 synthetic scenes from Replica\cite{straub2019replica} and qualitatively on 6 scenes from ScanNet\cite{dai2017scannet}.

\noindent\textbf{Metrics.} 
We employ several metrics to evaluate the reconstruction quality in our study. These include \textcolor{myBlue}{the peak signal-to-noise ratio (PSNR)}, Depth-L1 (on 2D depth maps), \textcolor{myBlue}{Structural Similarity} (SSIM\cite{wang2004image}), and \textcolor{myBlue}{Learned Perceptual Image Patch Similarity} (LPIPS\cite{zhang2018unreasonable}). Additionally, we assess the accuracy of camera pose estimation using the average absolute trajectory error (ATE RMSE \cite{sturm2012benchmark}). To evaluate the performance of semantic segmentation, we calculate the mIoU (mean Intersection over Union) score.

\noindent\textbf{Baselines.} 
We compare the tracking and mapping with state-of-the-art methods NICE-SLAM\cite{zhu2022nice}, Co-SLAM\cite{wang2023co}, ESLAM\cite{johari2023eslam}, and SplaTAM\cite{keetha2023splatam}. For semantic segmentation accuracy, we compare with NIDS-SLAM\cite{haghighi2023neural}, DNS-SLAM\cite{li2023dns}, and SNI-SLAM\cite{zhu2023sni}.

\noindent\textbf{Implementation Details.}
We conducted experiments using \textcolor{myBlue}{a single NVIDIA RTX 4090 and an Intel Xeon Platinum 8358P}, validating on the REPLICA dataset with the mapping iteration set to 40, tracking iteration set to 60, and SCFF iteration set to 50. After obtaining 384 feature channels through the DINO model, we derived 64-dimensional fused features by applying 2D convolutions separately to the Spatial Features. Finally, we obtained three-dimensional features by passing them through an encoder and embedding them into the gaussians parameters.
We use a learning rate of 0.005 and 0.001 respectively for all learnable parameters on Replica and ScanNet datasets. For camera poses, we only employ a learning rate of 0.0005 in tracking.

\begin{figure}
    \begin{center}
        \includegraphics[width=1\linewidth]{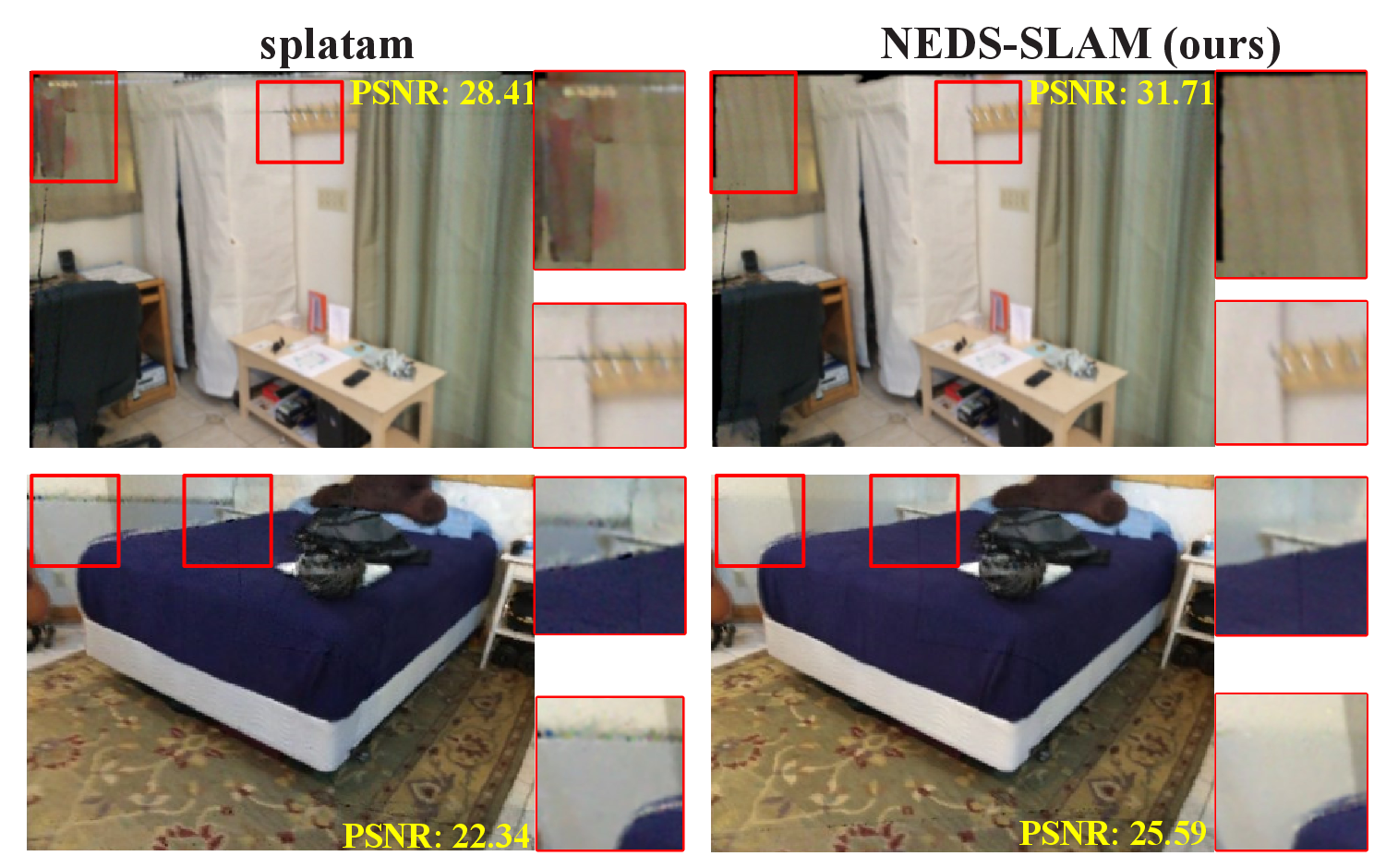}
    \end{center}
    \caption{
        The comparison validated the effectiveness of \textcolor{myBlue}{VCVP method}. Showing that NEDS-SLAM achieves better reconstruction results in details compared to Splatam. }
    \label{fig:expshow}
\end{figure}

\subsection{Experiment result}
Quantitative measures of reconstruction quality using the Replica dataset are presented in Table \ref{tab:exp1}. 
\textcolor{myBlue}
{The experiments on the ScanNet dataset can be found in Table \ref{tab:exp_scannet}. The data shows that our method achieves the highest camera pose tracking accuracy.
}
Our method demonstrates competitive performance when compared to other approaches.
As shown in Fig \ref{fig:expshow}, due to the \textcolor{myBlue}{VCVP} method removing \textcolor{myBlue}{geometrically unstable} gaussians, our approach is able to preserve more details.

The NEDS-SLAM, built upon the foundation of 3DGS, achieves accurate camera localization and semantic reconstruction simultaneously. Table \ref{tab:semantic_exp} provides a comparison between our method and other neural Implicit approaches in terms of semantic reconstruction performance.

Due to the precise representation of object edges offered by the \textcolor{myBlue}{3DGS}, \textcolor{myBlue}{our methods} bring about significant improvements in semantic reconstruction.
\textcolor{myBlue}
{
    Other methods have not considered the issue of spatially inconsistent semantic estimation by pre-trained semantic segmentation models on consecutive RGBD frame inputs. Therefore, for the sake of fair performance comparison in Table \ref{tab:semantic_exp}, we used the ground truth per-pixel semantic class labels as input. 
    More detailed experiments on the SCFF module are conducted in Table \ref{tab:scff_ablation} in Section \ref{sec:ablation}.
}

\begin{table}
    \centering 
    \caption{Comparison experiment on the mIoU metric}
    \begin{tabular}{c:c:ccc} 
    \toprule
    \multicolumn{1}{c}{Methods}                         & \multicolumn{1}{c}{AVG.mIoU[\%] $\uparrow$} & Room0          & Room1          & Office0         \\ 
    \midrule
    NIDS-SLAM\cite{haghighi2023neural} & 82.37                                       & 82.45          & 84.08          & 85.94           \\
    DNS-SLAM\cite{li2023dns}           & 84.77                                       & 88.32          & 84.90          & 84.66           \\
    SNI-SLAM\cite{zhu2023sni}          & 87.41                                       & 88.42          & 87.43          & 87.63           \\
    Ours                                                & \textbf{90.78}                              & \textbf{90.73} & \textbf{91.20} & \textbf{90.42}  \\
    \bottomrule
    \end{tabular}
    \label{tab:semantic_exp}
    \end{table}

When testing the \textcolor{myBlue}{Mask2Former} model on the replica room0 scene, as shown in Fig \ref{fig:fig4}, there are noticeable inconsistencies in the predictions for the floor and chairs. This affects the semantic reconstruction quality.
As shown in Fig \ref{fig:fig4}, NEDS-SLAM effectively filters out the negative impact of spatial semantic inconsistencies, generating robust semantic estimates and providing more accurate semantic reconstruction.

\begin{figure}
    \begin{center}
        \includegraphics[width=1\linewidth]{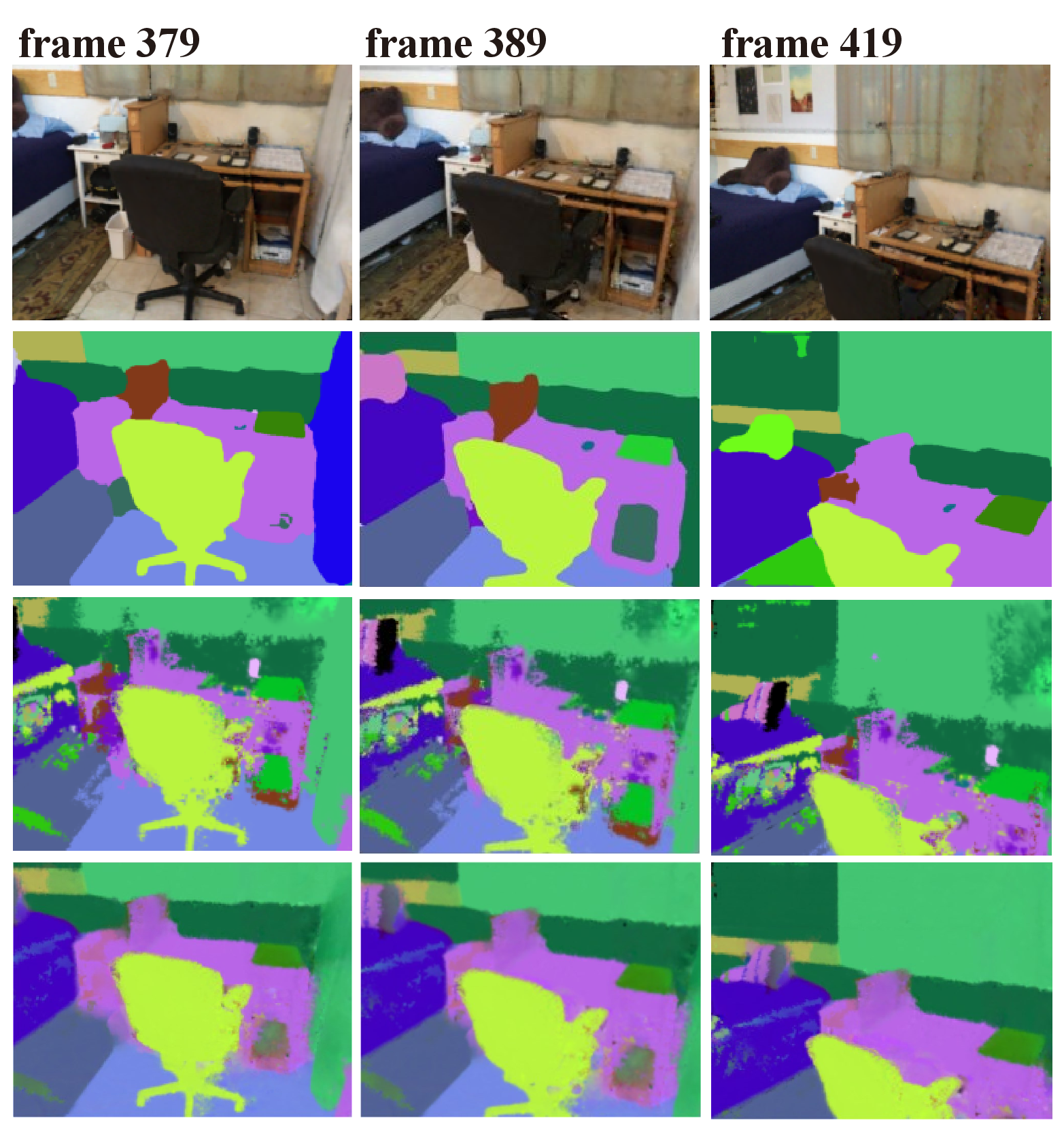}
    \end{center}
    \caption{
        The validation results on the Scannet scene0000\_00 dataset. The first row indicates the RGB reconstruction results of NEDS-SLAM, the second row indicates the semantic features predicted by M2F, the third row is the semantic reconstruction results without the Spatially Consistent Feature Fusion (SCFF) module, and the fourth row is the results with the SCFF module.
        }
    \label{fig:fig5}
\end{figure}

\subsection{Ablation Study}
\label{sec:ablation}

\textcolor{myBlue}{
\noindent\textbf{Effectiveness of VCVP Module.}}

\textcolor{myBlue}{
The VCVP method involves two subproblems: (1) determining the number of virtual camera views to generate and how to generate them, and (2) deciding on which frame or frames to perform VCVP operation. The solutions to these subproblems will impact the computational costs of the VCVP modules.
In Table \ref{tab:vcvp_ablation}, the data in the third and fourth rows labeled 'A/B/C' indicates that we used three configurations for the calculations. Configuration A and B represent generating two virtual views in the horizontal and vertical directions, respectively. Configuration C represents generating four virtual views simultaneously in both horizontal and vertical directions.
}

\textcolor{myBlue}{
The VCVP module significantly enhances scene modeling accuracy and camera pose tracking precision. Increasing the number of virtual camera views and their usage within the keyframes window can achieve the best camera localization accuracy, but this also increases computational overhead. In our most extreme test case, VCVP detection was performed on 10 keyframes during each mapping iteration, with four virtual camera viewpoints rendered for each detection, resulting in nearly a 50\% improvement in pose accuracy
}

\textcolor{myBlue}{
    We conducted another experiment comparing our method to the density control method (denote as DC method) from the original 3DGS paper, as in Table.\ref{tab:compare2originalGS}. In experiments on three scenes from the TUM RGBD dataset, the DC method achieved ATE RMSE values of 3.62, 1.41, and 6.63, which are higher than those of GS-SLAM, which also uses partial DC operations (3.3, 1.3, and 6.6 respectively). Our VCVP method demonstrated even higher performance.
}

\begin{table}\textcolor{myBlue}{
    \centering
    \resizebox{\linewidth}{!}{%
    \begin{threeparttable}  
    \caption{Ablation experiments on the VCVP module conducted in replica room0.}
    \label{tab:vcvp_ablation}
    \begin{tabular}{lcccc} 
    \toprule
    Settings         & \begin{tabular}[c]{@{}c@{}}ATE \\RMSE $\downarrow$\end{tabular} & \begin{tabular}[c]{@{}c@{}}AVG \\SSIM $\uparrow$\end{tabular} & \begin{tabular}[c]{@{}c@{}}Scene \\Embedding $\downarrow$\end{tabular} & \begin{tabular}[c]{@{}c@{}}Mapping\\/iteration $\downarrow$\end{tabular}  \\ 
    \midrule
    Base\tnote{1}             & 0.42                                                            & 0.90                                                          & 100.16 MB                                                              & 14 ms                                                                        \\
    Base + VCVP\_w5\tnote{*}  & 0.30/0.34/0.28                                                  & 0.91/0.93/0.96                                                & 90.45 MB                                                              & 16/16/20 ms                                                                 \\
    Base + VCVP\_w10 & 0.26/0.27/0.22                                                  & 0.92/0.92/0.97                                                & 88.93 MB                                                               & 18/18/26 ms                                                                 \\
    Base + RCVP\tnote{2}      & 0.36                                                            & 0.95                                                          & 95.27 MB                                                              & 20 ms                                                                        \\ 
    \hdashline
    SplaTAM\cite{keetha2023splatam}          & 0.36                                                            & 0.98                                                          & 100.00 MB                                                              & 24 ms                                                                        \\
    Co-SLAM\cite{wang2023co}          & 0.97                                                            & 0.91                                                          & -                                                                      & 13 ms                                                                        \\
    NICE-SLAM\cite{zhu2022nice}        & 0.99                                                            & 0.69                                                          & 48.48 MB                                                               & 66 ms                                                                        \\
    \bottomrule
    \end{tabular}
    \begin{tablenotes} 
        \footnotesize
        \item[1] \textbf{Base} refers to the configuration without SCFF, without lightweight encoder, and without VCVP, implementing only the 3DGS dense SLAM functionality.
        \item[*] \textbf{VCVP\_w5(10)} indicates selecting 5(10) frames from the current keyframes window for VCVP operations.
        \item[2] \textbf{RCVP} involves using real camera views for consistency checks and removing outlier gaussians.
    \end{tablenotes}
    \end{threeparttable}}
}   
\end{table}

\begin{table}\textcolor{myBlue}{
    \centering
    \resizebox{\linewidth}{!}{%
    \begin{threeparttable}
    \caption{Ablation Study of the SCFF Module on ScanNet dataset.}
    \label{tab:scff_ablation}
    \begin{tabular}{lccc} 
    \toprule
    Settings                                                                   & mIoU~ $\uparrow$   & \begin{tabular}[c]{@{}c@{}}Mapping\\/iteration~$\downarrow$\end{tabular} & \begin{tabular}[c]{@{}c@{}}Scene \\Embedding~$\downarrow$\end{tabular}  \\ 
    \midrule
    Base\_S                                                                    & 26.52\% & 28 ms                                                        & 123.08 MB                                                   \\ 
    \hdashline[1pt/1pt]
    \begin{tabular}[c]{@{}l@{}}Base\_S + SCFF\_wo\_SFE\end{tabular}          & 30.24\% & 86 ms                                                        & 405.64 MB                                                   \\ 
    \hdashline[1pt/1pt]
    \begin{tabular}[c]{@{}l@{}}Base\_S + SCFF\_w\_SFE\end{tabular}           & 42.18\% & 86 ms                                                        & 410.38 MB                                                   \\ 
    \hdashline[1pt/1pt]
    \begin{tabular}[c]{@{}l@{}}Base\_S + SCFF\_w\_SFE + encoder\end{tabular} & 40.81\% & 35 ms                                                        & 141.93 MB                                                   \\
    \toprule
    \end{tabular}
    \end{threeparttable}}
}
\end{table}


\begin{table}
    \textcolor{myBlue}{
    \centering
    \caption{Runtime performance comparison of NEDS-SLAM on two different hardware platforms.}
    \label{tab:hardware_settings}
    \resizebox{\linewidth}{!}{%
    \begin{tabular}{lcccc} 
    \toprule
    \multirow{2}{*}{\begin{tabular}[c]{@{}l@{}}Hardware\\Settings\end{tabular}} & \multicolumn{2}{c}{\# replica room0}                                                                                                                               & \multicolumn{2}{c}{\# TUM RGBD fr1/desk}                                                                                                                            \\ 
    \cmidrule(l){2-5}
                                                                                & \multicolumn{1}{l}{\begin{tabular}[c]{@{}l@{}}Tracking/it $\downarrow$\end{tabular}} & \multicolumn{1}{l}{\begin{tabular}[c]{@{}l@{}}Mapping/it $\downarrow$\end{tabular}} & \multicolumn{1}{l}{\begin{tabular}[c]{@{}l@{}}Tracking/it $\downarrow$\end{tabular}} & \multicolumn{1}{l}{\begin{tabular}[c]{@{}l@{}}Mapping/it $\downarrow$\end{tabular}}  \\ 
    \midrule
    Platform A                                                                  & 28 ms                                                                            & 35 ms                                                                           & 26 ms                                                                            & 34 ms                                                                            \\
    Platform B                                                                  & 42 ms                                                                            & 76 ms                                                                           & 42 ms                                                                            & 75 ms                                                                            \\
    \bottomrule
    \end{tabular}}
    }
    
    \end{table}


\begin{table}
    \centering
    \caption{\textcolor{myBlue}{Comparison of the VCVP module with the original density control method on TUM-RGBD dataset.}}
    \resizebox{\linewidth}{!}{%
    \textcolor{myBlue}{
    \begin{tabular}{lcccc} 
    \toprule
    \multirow{2}{*}{DATASETS} & \multicolumn{2}{c}{with VCVP} & \multicolumn{2}{c}{Original density control method
      as in \cite{kerbl20233d}}  \\ 
    \cmidrule(lr){2-5}
                              & ATE
      RMSE $\downarrow$ & AVG
      SSIM $\uparrow$      & ATE
      RMSE$\downarrow$ & AVG
      SSIM $\uparrow$                                      \\ 
    \midrule
    Fr1/desk1                 & 3.30       & 0.91             & 3.62       & 0.93                                             \\
    Fr2/xyz                   & 1.13       & 0.95             & 1.41       & 0.95                                             \\
    Fr3/off                   & 4.94       & 0.90             & 6.63       & 0.92                                             \\
    \bottomrule
    \end{tabular}
    }}
    
    \label{tab:compare2originalGS}
    \end{table}


    

\begin{table}
    \textcolor{myBlue}{
    \centering
    \caption{Verification of the effectiveness of the SCFF module}
    \resizebox{\linewidth}{!}{%
    \begin{tabular}{lcccc} 
    \toprule
    Model Settings & M2F\cite{cheng2021mask2former}     & M2F+SCFF & MRCNN\cite{matterport_maskrcnn_2017}   & MRCNN+SCFF  \\ 
    \midrule
    AVG mIoU       & 25.89\% & 36.25\%  & 24.34\% & 34.07\%     \\
    \bottomrule
    \end{tabular}}
    \label{tab:scff_sota}
    }
\end{table}
\textcolor{myBlue}{
\noindent\textbf{Effectiveness of SCFF Module.}}

Following SGS-SLAM, we directly incorporated semantic parameters into the \textcolor{myBlue}{3D gaussians} by calling a pre-trained M2F segmentation model \cite{cheng2021mask2former} on each RGB frame. As shown in the \textcolor{myBlue}{third row in} Fig \ref{fig:fig4}, corresponding to the \textcolor{myBlue}{Base\_S settings in} Table \ref{tab:scff_ablation}.
\textcolor{myBlue}{
SCFF\_wo\_SFE represents configurations includes the SCFF module, but does not use SFE.
For the ScanNet scene0000 dataset, the M2F model achieved a semantic segmentation mIoU of 52.4. Using M2F for segmentation head gave an average mIoU of 26.52, serving as the baseline.}


As can be seen in Fig \ref{fig:fig5}, the semantic features calculated by the M2F model were inconsistent (such as the partitions and books on the table). After processing with the SCFF module, the inconsistencies were resolved and NEDS-SLAM output a more complete semantic reconstruction.
\textcolor{myBlue}{The SCFF module filters out unstable semantic estimations between frames, resulting in more accurate semantic reconstruction. Our designed SCFF features a lightweight network structure, which does not significantly increase inference time.
In fact, the time consumption in the SLAM process (Mapping/Iteration in the table) mainly arises from optimizing a large number of 3D gaussians. Therefore, our specially designed encoder compresses the semantic features and embeds them into the gaussian parameters, reducing the number of parameters and thereby increasing the mapping speed.
}

\textcolor{myBlue}{
\noindent\textbf{Runtime Comparison.}}

\textcolor{myBlue}{
As shown in the last column of Table \ref{tab:hardware_settings}, our lightweight configuration of NEDS-SLAM achieves faster mapping speeds than SplaTAM while maintaining more accurate camera pose tracking precision. With higher configurations, NEDS-SLAM offers better performance, though the computation speed decreases.
We ran NEDS-SLAM on different hardware platforms and datasets. Platform A is as in section \ref{sec:experiment_setup}. Platform B consists of an Intel i9-13900K and a single NVIDIA RTX 4060Ti. The results show that both hardware platforms achieve similar camera pose tracking accuracy and reconstruction accuracy with NEDS-SLAM, but the model takes more time to run on Platform B compared to Platform A.
}

%% file: secs/conclusion.tex
\section{Conclusion and Limitations}
\label{sec:limitation_conclusion}
The proposed NEDS-SLAM is an end-to-end semantic SLAM system based on 3DGS. By integrating a Spatially Consistent feature fusion model, NEDS-SLAM effectively addresses the challenges of robustly estimating semantic labels with pre-trained models, significantly enhancing semantic reconstruction performance.
The Virtual Camera View Pruning method uses differentiable Gaussian splatting for quick and realistic novel view synthesis. It removes outlier gaussians during SLAM, significantly improving the reconstruction quality of neural radiance fields.

The experiment with public datasets confirmed NEDS-SLAM's effectiveness but revealed some shortcomings. 
The virtual camera view pruning method improves mapping speed by removing more gaussians. However, increasing the frequency of VCVP usage also raises computational load, indicating room for further optimization.
Future plans include optimizing and incorporating semantic reconstruction for dynamic scenes.